\documentclass[conference]{IEEEtran}
\usepackage{balance}

\usepackage{multirow}
\usepackage{tikz}
\usetikzlibrary{shapes,decorations}

\newcommand{\shapelabel}[3]{
    \begin{tikzpicture}[baseline={([yshift=-0.8ex]current bounding box.center)}]
    \protect\node[#1,fill=#2,draw=black,scale=.8]{\color{white}\textbf{#3}};
    \end{tikzpicture}
}

\tikzset{
    shapelabel/.style n args={2}{
        #1,
        draw=black,
        fill=#2,
        baseline={([yshift=-0.8ex]current bounding box.center)},
        text=white
    }
}

\usepackage{cite}
\usepackage{hyperref}

\usepackage{xcolor}
\usepackage{xfrac}
\usepackage{caption}
\usepackage{subcaption}
\usepackage{siunitx}
\usepackage{amsmath}
\usepackage{amssymb}
\usepackage{amsfonts}
\usepackage{graphicx}
\usepackage{textcomp}
\usepackage{xcolor}
\usepackage{multirow}
\usepackage{soul}
\usepackage{algorithm}
\usepackage{algpseudocode}
\usepackage{colortbl}
\usepackage{bm}
\usepackage{float}
\usepackage{filecontents}
\usepackage{graphics}
\usepackage{csvsimple}

\begin{document}
%
\title{Multi-Objective Neural Architecture Search for In-Memory Computing}


\author{\IEEEauthorblockN{Md Hasibul Amin, Mohammadreza Mohammadi, Ramtin Zand}
\IEEEauthorblockA{Department of Computer Science and Engineering, University of South Carolina, Columbia, SC 29208, USA\\
e-mail: ma77@email.sc.edu, mohammm@email.sc.edu, ramtin@cse.sc.edu
}
}

\maketitle

\begin{abstract}

In this work, we employ neural architecture search (NAS) to enhance the efficiency of deploying diverse machine learning (ML) tasks on in-memory computing (IMC) architectures. Initially, we design three fundamental components inspired by the convolutional layers found in VGG and ResNet models. Subsequently, we utilize Bayesian optimization to construct a convolutional neural network (CNN) model with adaptable depths, employing these components. Through the Bayesian search algorithm, we explore a vast search space comprising over 640 million network configurations to identify the optimal solution, considering various multi-objective cost functions like accuracy/latency and accuracy/energy. Our evaluation of this NAS approach for IMC architecture deployment spans three distinct image classification datasets, demonstrating the effectiveness of our method in achieving a balanced solution characterized by high accuracy and reduced latency and energy consumption.

\end{abstract}



%
\IEEEpeerreviewmaketitle

\section{Introduction}

In-memory computing (IMC) architectures, also referred to as processing-in-memory (PIM) or compute-in-memory (CIM), have emerged as promising alternatives to traditional von Neumann-based machine learning (ML) hardware \cite{IMCSurvey,TPU-IMAC}. These architectures capitalize on characteristics such as massive parallelism, analog computation, and executing computations directly where data is stored, resulting in notable performance enhancements \cite{PUMA,prime,ISAAC}. The backbone of most IMC architectures lies in memristive crossbar arrays, which leverage resistive memory technologies like resistive random-access memory (RRAM) \cite{RRAM} and magnetoresistive random-access memory (MRAM) \cite{zand2018fundamentals}. These arrays enable matrix-vector multiplication operations in the analog domain, using fundamental circuit principles such as Ohm’s Law and Kirchoff’s Current Law \cite{iCASISVLSI22,dpengine}.

Despite the aforementioned advancements, prior research has shown that deploying ML models that are pre-trained and optimized using digital von Neumann architectures, such as CPU and GPU, on an analog IMC architecture 
does not consistently yield comparable performance \cite{xbarpartition}. Several factors contribute to this, including limited numerical precision of memristive devices \cite{mixedIMC}, as well as circuit imperfections such as the interconnect parasitic \cite{parasiticsiCAS}, and device-to-device and cycle-to-cycle variations \cite{variation}. The \textit{in-circuit training} \cite{prezioso2015training} is introduced as a mechanism to address the deployment challenges by bringing the IMC circuits within the loop of training and allowing the ML models to learn the circuit imperfections. However, this method may face limited applicability due to the endurance constraints of memristive devices \cite{Endurance1}. These devices can lose their storage capability after a certain number of repeated write operations.





Another strategy involves utilizing IMC hardware simulators like Neurosim \cite{neurosim}, MNSIM \cite{MNSIM2}, and IMAC-Sim \cite{imac-sim} to emulate the characteristics of IMC circuits and the mapping strategies for deploying ML models on an IMC chip. This method, which we termed \textit{pre-silicon optimization}
, involves adjusting circuit and device level parameters to achieve specific objectives. One advantage of this approach is its holistic exploration of both IMC circuit and neural architecture parameters. However, pre-silicon optimization methods \cite{gibbon, nacim, nas4rram, uae} are primarily intended for chip manufacturers to make early design decisions before chip fabrication.

Our paper focuses on another mechanism, termed \textit{post-silicon optimization}\footnote{The term ``post-silicon'' does not imply the necessity of physical hardware for optimization. Instead, it emphasizes optimizing parameters that can be adjusted after hardware fabrication.}
, which integrates hardware deployment into the optimization loop but only adjusts neural architecture parameters related to the ML model while keeping hardware parameters fixed. This approach utilizes hardware measurements such as accuracy, latency, and energy to formulate a multi-objective fitness function (FF). The neural architecture search (NAS) algorithm leverages this fitness function to develop models that can perform effectively despite IMC circuit limitations, without modifying the circuit itself. While our paper does not employ physical hardware implementation of IMC circuits, the proposed methodologies can be readily applied to optimize and deploy various ML models on IMC architectures.

\section{Analog IMC Architecture Background}

\begin{figure*}
    \centering
    \includegraphics[width=7in]{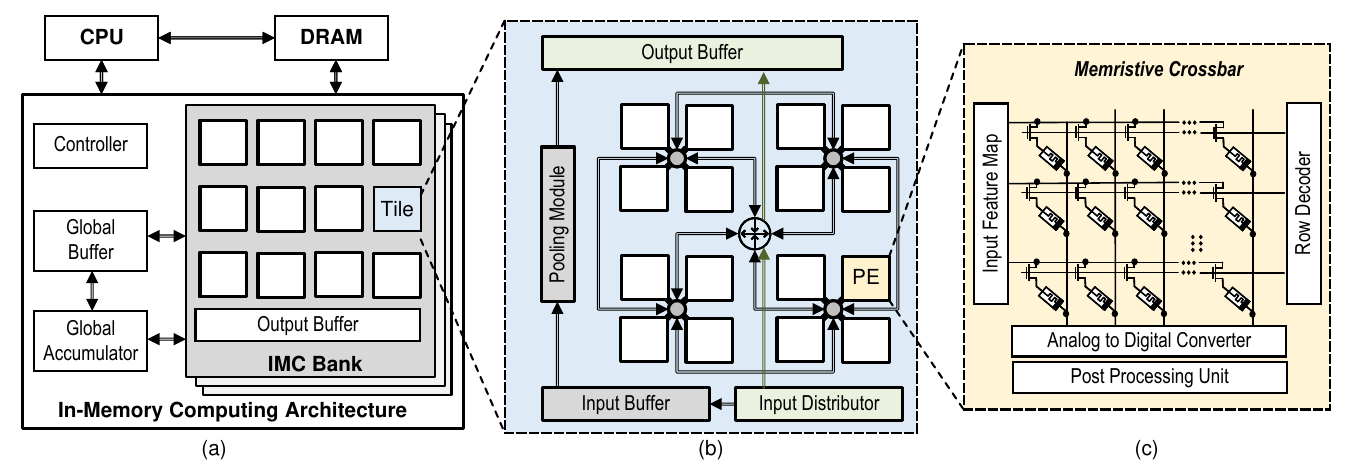}
    \caption{(a) The analog IMC architecture with multiple IMC banks each of which includes several interconnected IMC tiles \cite{MNSIM2}. (b) The IMC tile consists of a network of tightly coupled processing elements (PEs). (c) The structure of the IMC processing element that includes memristive crossbars in its core.}
    \label{fig:arch}
\end{figure*}

\begin{figure}
    \centering
    \includegraphics[width=3.4in]{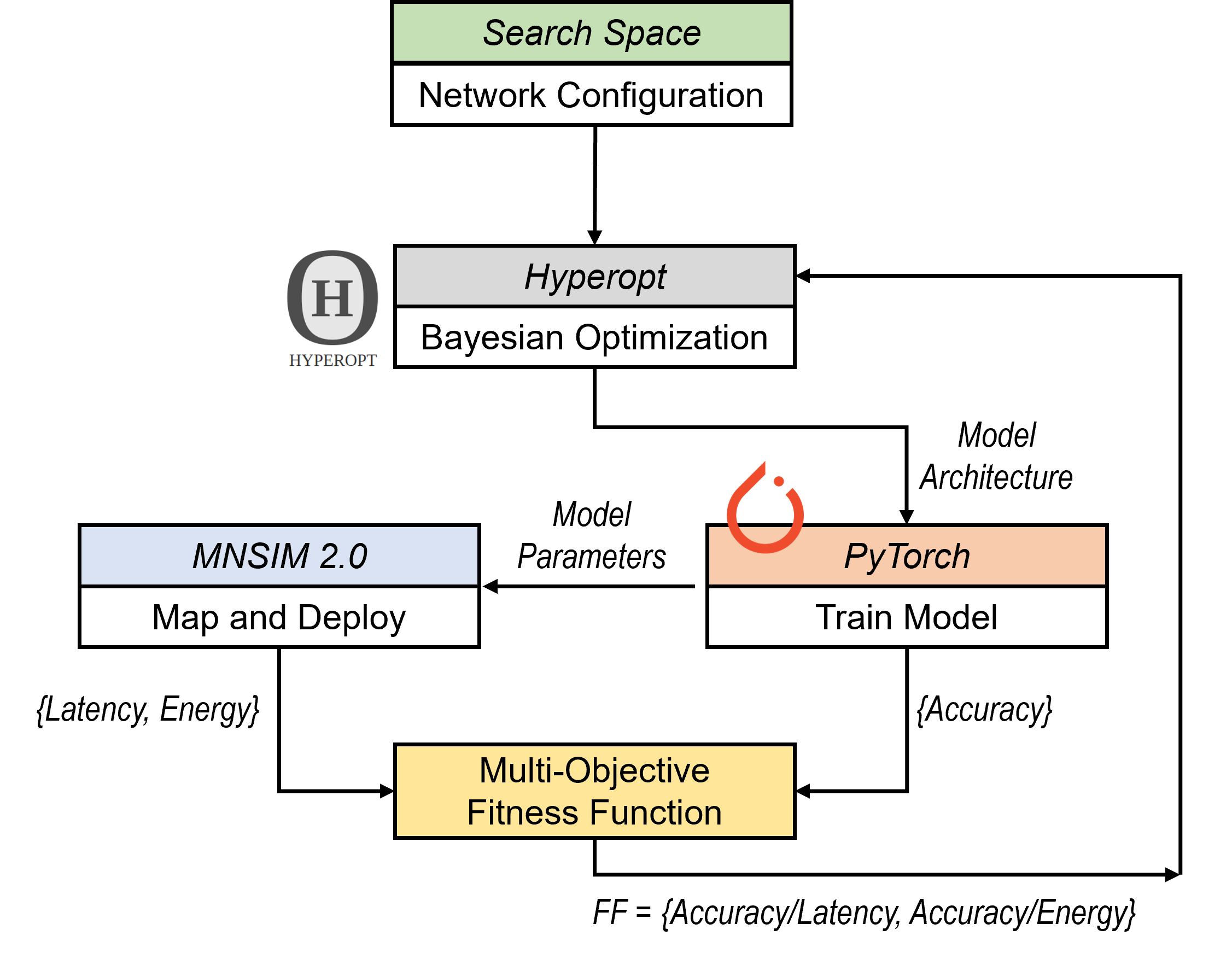}
    \caption{The proposed NAS methodology for multi-objective optimization of ML workloads deployed on analog IMC architectures.}
    \label{fig:methodology}
\end{figure}

Figure \ref{fig:arch} demonstrates the hierarchical structure of the In-Memory Computing (IMC) architecture \cite{MNSIM2}. The IMC architecture consists of multiple IMC banks, the global buffer, the global accumulator, and the IMC controller as shown in Fig. \ref{fig:arch}a. The IMC controller receives the status updates from the CPU and controls the data movement between DRAM and the IMC banks. Global buffer and global accumulator are necessary for the elementwise-sum operation to implement the skip-connections in ML models like ResNet \cite{he2016deep}. Inside an IMC bank, arrays of tiles are connected using a network-on-chip structure. Inside a tile, there are multiple processing elements (PE) and a pooling module along with input and output buffers as shown in Fig. \ref{fig:arch}b. The PE blocks contain crossbars of resistive memory devices such as RRAM and the peripheral circuits. Peripheral circuits vary based on analog and digital IMC. Figure \ref{fig:arch}c shows the internal structure of an analog PE. Analog IMC requires analog-to-digital converter (ADC) and digital-to-analog converter (DAC) blocks as peripheral circuits, while digital IMC requires digital computational units, shift-and-add circuits, and sense amplifiers. Crossbars perform matrix-vector multiplication (MVM) operations in either analog or digital domains. In the analog IMC, which is the architecture utilized in this paper, the weight kernels are expanded into a vector and loaded onto the columns of the crossbars. The input feature map is also expanded into a vector and provided as inputs to the crossbar. The crossbar performs the MVM operation using fundamental circuit principles in parallel and in $O(1)$ time complexity. 

\begin{figure*}
     \centering
     \captionsetup{font=footnotesize}
     \begin{subfigure}[b]{0.3\textwidth}
         \centering
         \includegraphics[width=\textwidth]{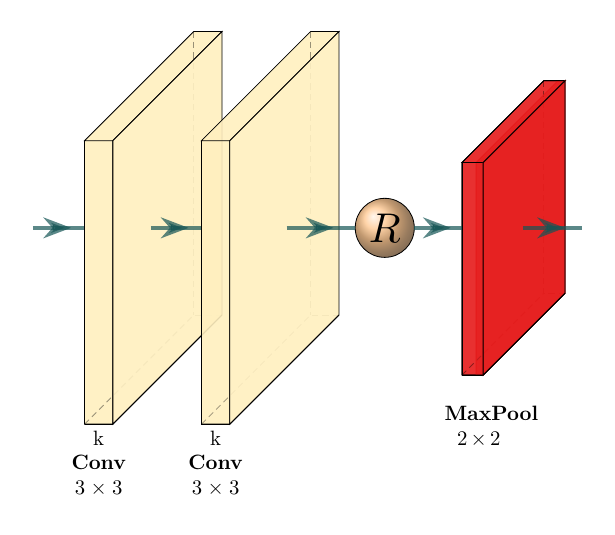}
         \caption{VGG block with MaxPooling}
         \label{fig:vgg}
     \end{subfigure}
     \hfill
     \begin{subfigure}[b]{0.25\textwidth}
         \centering
         \includegraphics[width=\textwidth]{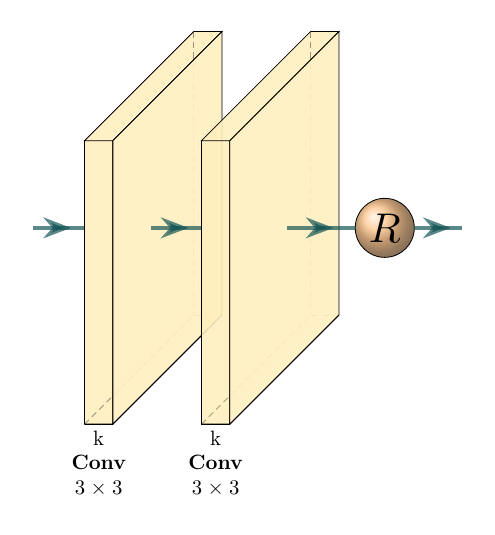}
         \caption{MVGG block without MaxPooling}
         \label{fig:mvgg}
     \end{subfigure}
     \hfill
     \begin{subfigure}[b]{0.35\textwidth}
         \centering
         \includegraphics[width=\textwidth]{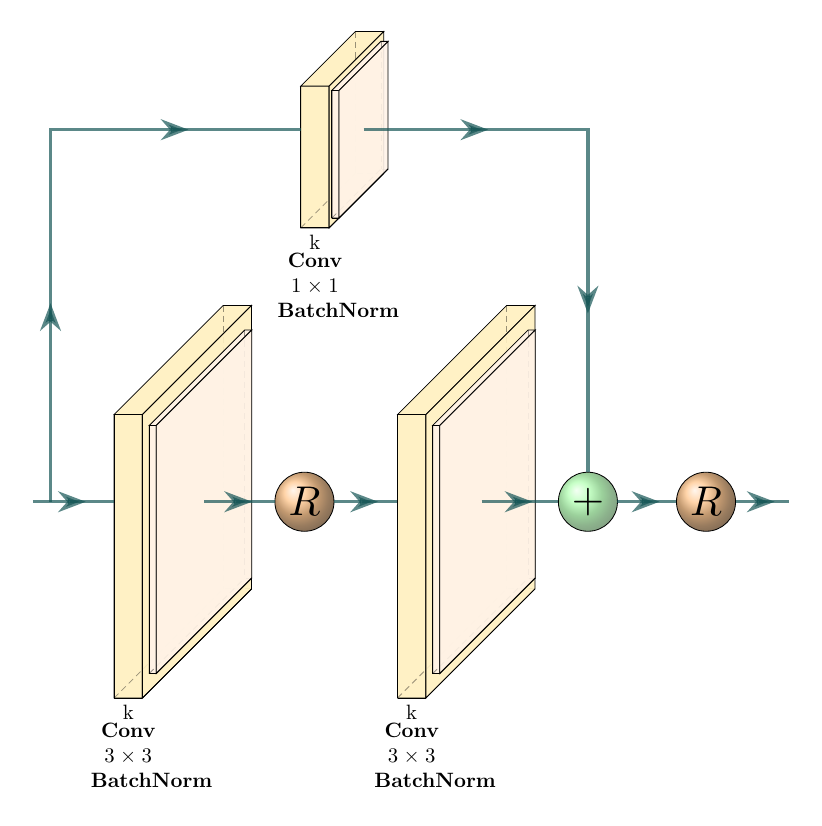}
         \caption{ResNet block (RES)}
         \label{fig:res}
     \end{subfigure}
        \caption{The three building blocks of the CNN architecture.}
        \label{fig:blocks}
\end{figure*}

\section{The Proposed NAS Methodology}

 

\begin{figure*}
    \centering
    \includegraphics[width=7in]{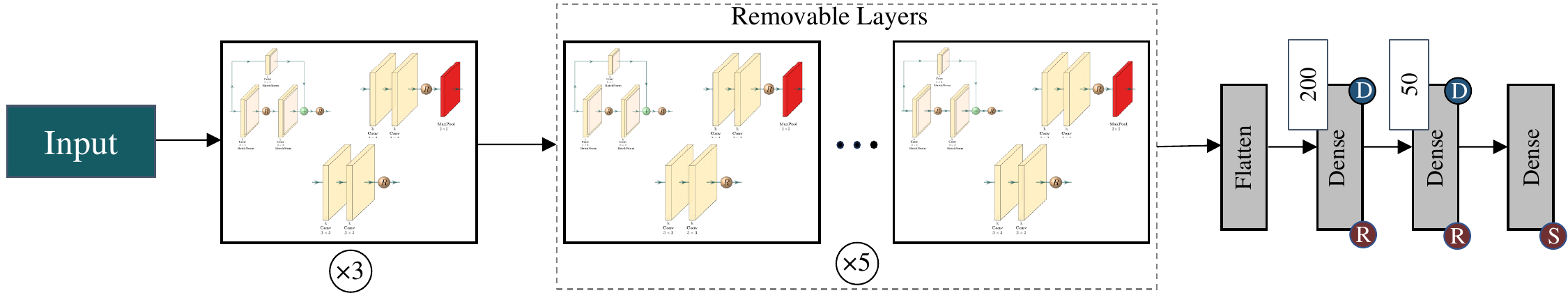}
    \caption{The network configuration. \shapelabel{circle}{black!60!red}{R} ReLU, \shapelabel{circle}{blue!60!green}{D} Dropout, \shapelabel{circle}{black!60!red}{S} Softmax.}
    \label{fig:arch}
\end{figure*}

In this paper, we present a novel neural architecture search (NAS) approach for multi-objective optimization of the ML workloads deployed on analog IMC architectures. The high-level description of the proposed methodology is illustrated in Fig. \ref{fig:methodology}. 
We employ the Hyperopt library \cite{bergstra2015hyperopt} to handle the NAS process. Hyperopt \cite{bergstra2015hyperopt} is a Python library designed for automated hyperparameter tuning, employing various sequential model-based optimization techniques, also known as Bayesian optimization. The search process is an iterative process that is repeated for a certain number of iterations set by the user, and the best model is selected based on a pre-defined fitness function (FF). 

We have integrated Hyperopt with PyTorch and MNSIM 2.0 \cite{MNSIM2} frameworks to perform the multi-objective optimization. In each iteration, Hyperopt automatically selects a model configuration from the search space and uses PyTorch and MNSIM to obtain the model accuracy as well as the IMC hardware performance metrics such as latency and energy. These metrics are combined to form multi-objective FF such as $\sfrac{accuracy}{latency}$ or $\sfrac{accuracy}{energy}$.


Before beginning the optimization, we should first create the search space. Here, we focus on convolutional neural networks (CNNs) as representative ML workloads. We design three distinct building blocks: (1) the VGG block with sub-sampling, (2) the modified VGG block (MVGG) without sub-sampling, and (3) the ResNet block (RES). As shown in Fig. \ref{fig:vgg}, the VGG block with sub-sampling consists of two consecutive convolutional layers with a kernel size of $3\times3$. The number of kernels ($k$) is determined during the optimization phase. Following these layers, a ReLU activation function and a $2\times2$ max-pooling layer are applied. In the MVGG block depicted in Fig. \ref{fig:mvgg}, we remove the sub-sampling block. This decision is made to leverage the inherent properties of the VGG block without downsizing the feature map dimensions within the network. The RES block, depicted in Fig. \ref{fig:res}, consists of two consecutive convolutional layers, each configured with $k$ kernels and a $3\times3$ kernel size. Following each convolutional layer, a batch normalization layer is applied, with a ReLU activation function positioned between them. On the residual connection side, there is a convolutional layer with $k$ kernels, each having a size of $1\times1$. Similar to VGG blocks, the number of kernels is determined during the optimization phase. The outputs from both the convolutional layers and the residual connection are combined, and then a ReLU function is applied.

The overall structure of the CNN models created in this paper is illustrated in Fig. \ref{fig:arch}, incorporating the aforementioned building blocks. We include several adjustable parameters to optimize the networks. As outlined in Table \ref{tab:params}, the parameters available for tuning are (1) the number of blocks ($Block$) can vary between 3 and 8, allowing for different network depths; (2) the type of each block ($BT$) can be any of the three building blocks, including VGG with sub-sampling ($VGG$), modified VGG without sub-sampling ($MVGG$), and ResNet block ($RES$); (3) the number of kernels ($K$) for each block ranging from 16 to 256. After the \textit{flatten} layer, the number of layers and neurons in the fully connected layers remain fixed. These configurable parameters collectively define a search space comprising over 641 million unique network configurations. 

\begin{table}[]
\centering
\caption{Network configuration settings}
\begin{tabular}{cccc}
\hline
Parameters & Description                     & Options              \\ \hline
$Block$      & No. of blocks in the network & 3-8            \\
$BT$         & The block type               & Res, VGG, MVGG     \\
$K$          & No. of kernels in each conv layer               & 16, 32, 64, 128, 256        \\ \hline
\end{tabular}
\label{tab:params}
\end{table}

\section{Results}

\begin{figure*}
     \centering
     \begin{subfigure}[b]{0.28\textwidth}
         \centering
         \includegraphics[width=\textwidth]{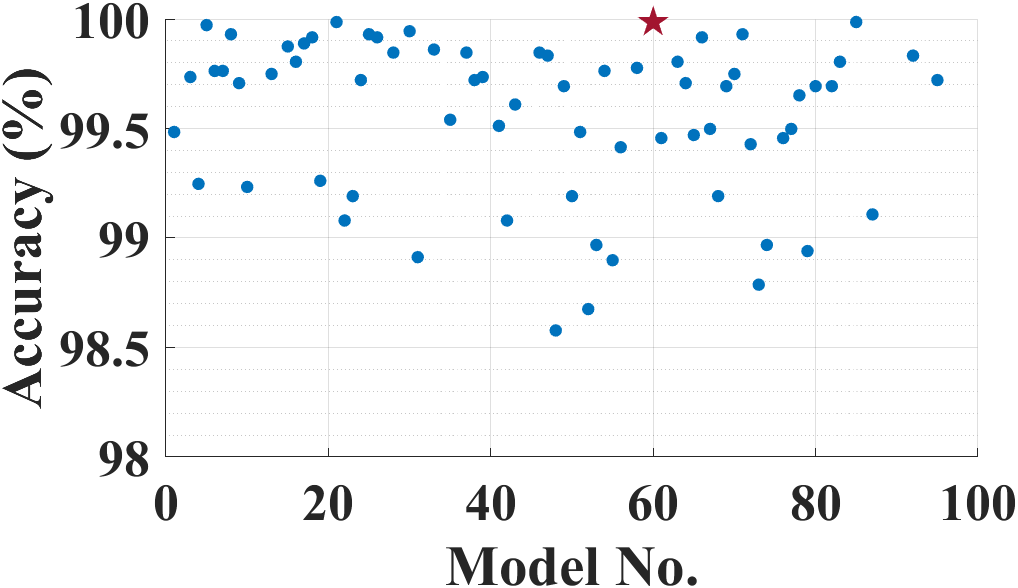}
         \caption{}
         \label{fig:asl_acc1}
     \end{subfigure}
     \hfill
     \begin{subfigure}[b]{0.28\textwidth}
         \centering
         \includegraphics[width=\textwidth]{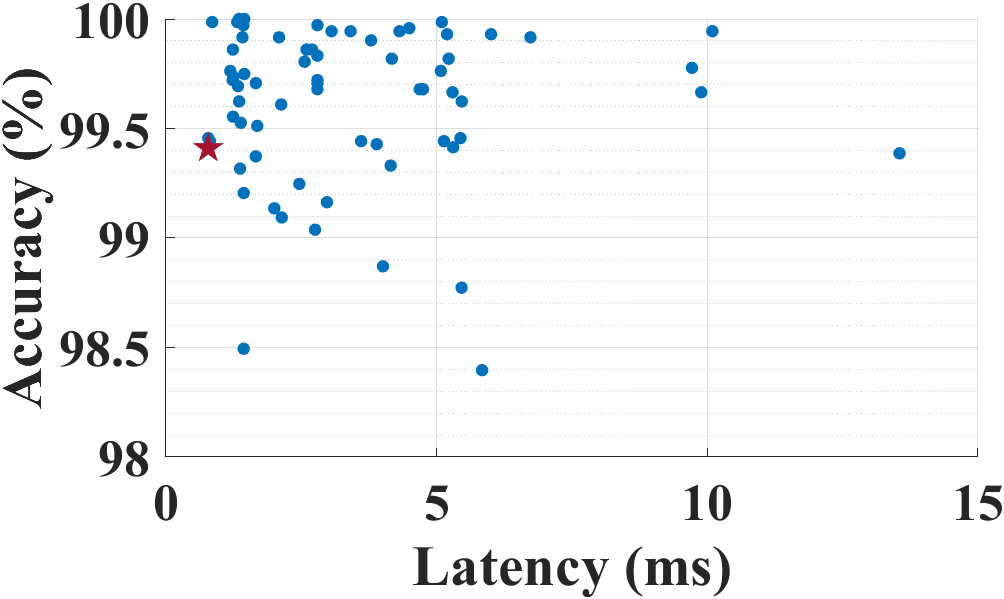}
         \caption{}
         \label{fig:asl_lt1}
     \end{subfigure}
     \hfill
     \begin{subfigure}[b]{0.28\textwidth}
         \centering
         \includegraphics[width=\textwidth]{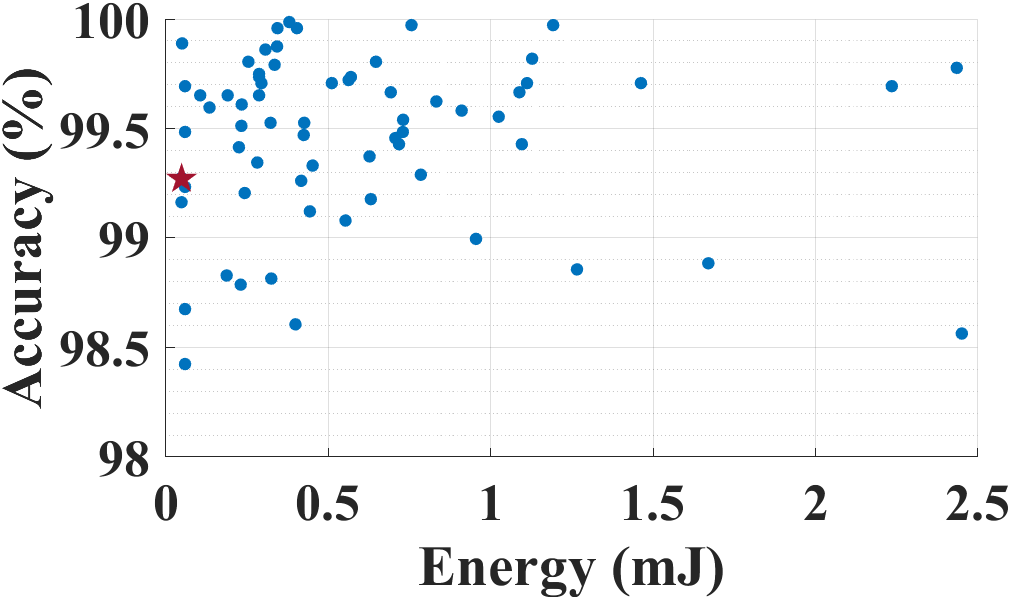}
         \caption{}
         \label{fig:asl_eng1}
     \end{subfigure}
        \caption{Distribution of NAS outputs for ASL dataset (The best model is marked by \textcolor{red}{$\star$}) (a) distribution of accuracy when $FF=accuracy$ (b) distribution of accuracy vs latency when $FF=\sfrac{accuracy}{latency}$ (c) distribution of accuracy vs energy when $FF=\sfrac{accuracy}{energy}$.}
        \label{fig:asl_scat}
\end{figure*}

\begin{figure*}
     \centering
     \begin{subfigure}[b]{0.28\textwidth}
         \centering
         \includegraphics[width=\textwidth]{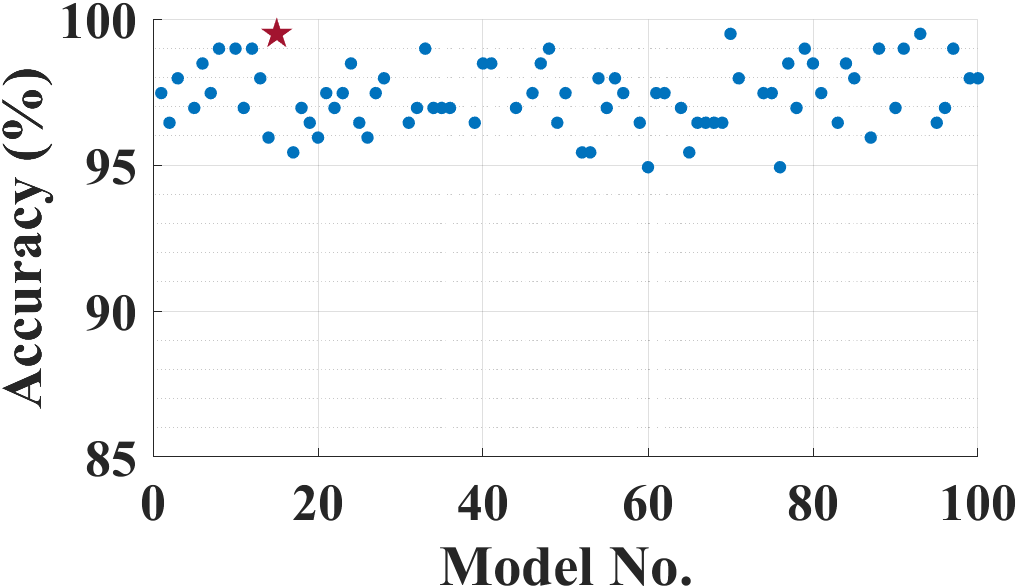}
         \caption{}
         \label{fig:ck_acc}
     \end{subfigure}
     \hfill
     \begin{subfigure}[b]{0.28\textwidth}
         \centering
         \includegraphics[width=\textwidth]{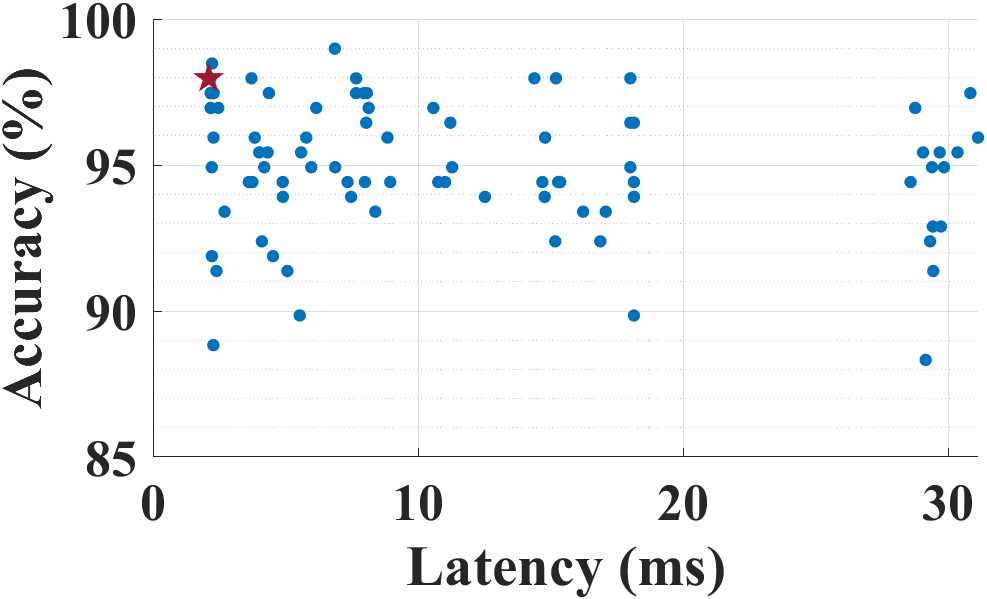}
         \caption{}
         \label{fig:ck_lt}
     \end{subfigure}
     \hfill
     \begin{subfigure}[b]{0.28\textwidth}
         \centering
         \includegraphics[width=\textwidth]{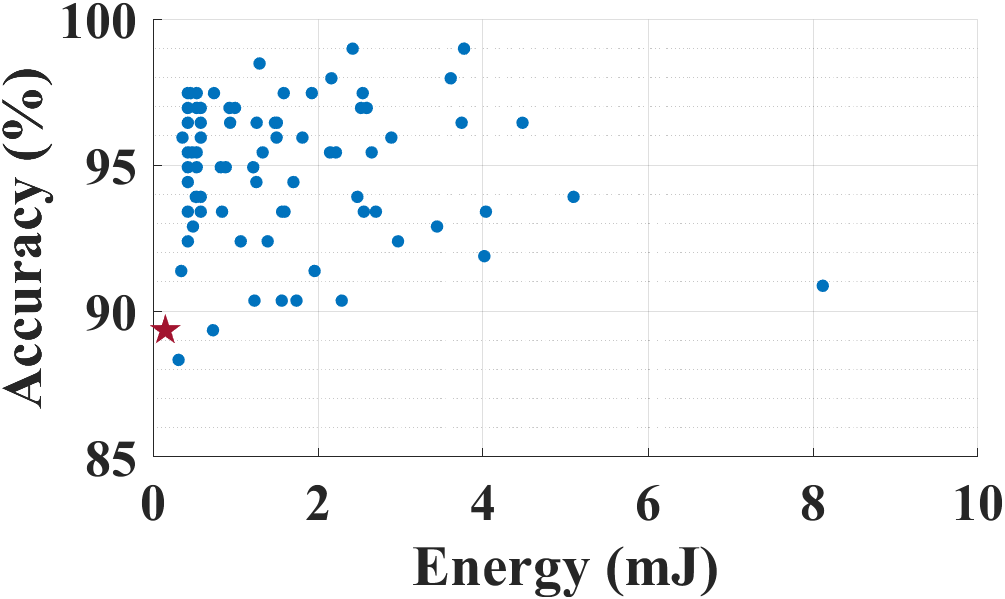}
         \caption{}
         \label{fig:ck_eng}
     \end{subfigure}
        \caption{Distribution of NAS outputs for CK+ dataset (The best model is marked by \textcolor{red}{$\star$}) (a) distribution of accuracy when $FF=accuracy$ (b) distribution of accuracy vs latency when $FF=\sfrac{accuracy}{latency}$ (c) distribution of accuracy vs energy when $FF=\sfrac{accuracy}{energy}$.}
        \label{fig:ck_scat}
\end{figure*}

\begin{figure*}
     \centering
     \begin{subfigure}[b]{0.28\textwidth}
         \centering
         \includegraphics[width=\textwidth]{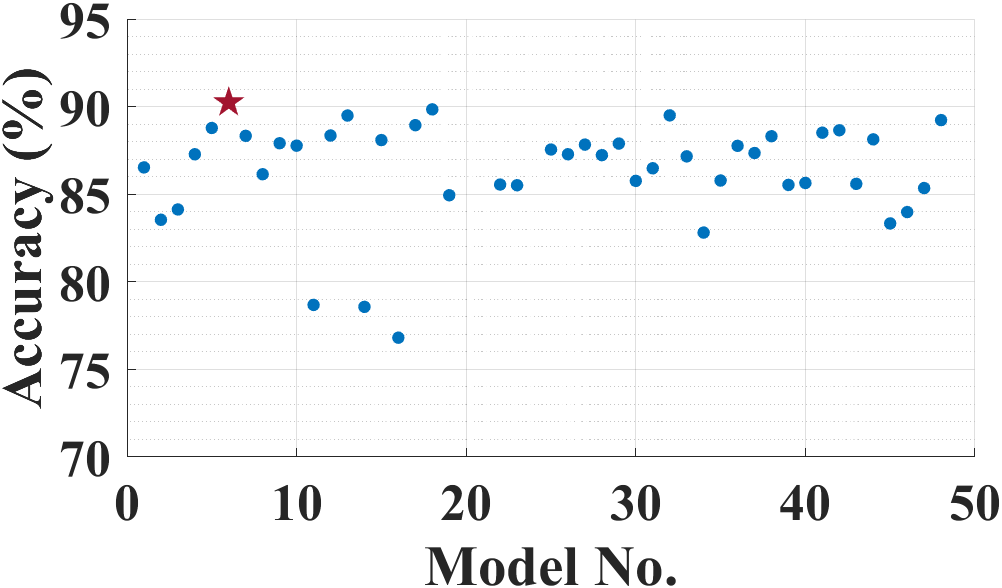}
         \caption{}
         \label{fig:cifar_acc}
     \end{subfigure}
     \hfill
     \begin{subfigure}[b]{0.28\textwidth}
         \centering
         \includegraphics[width=\textwidth]{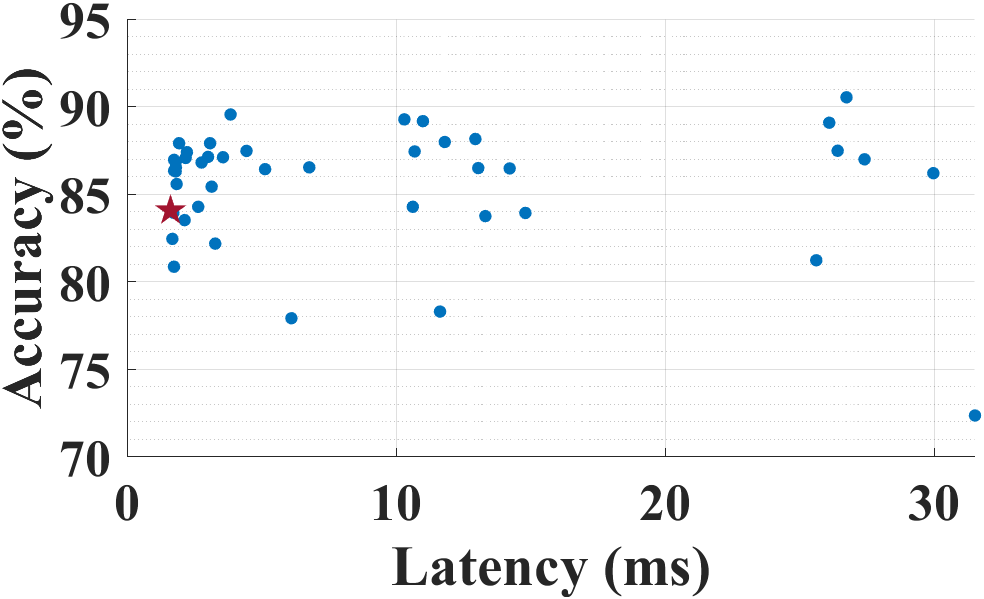}
         \caption{}
         \label{fig:cifar_lt}
     \end{subfigure}
     \hfill
     \begin{subfigure}[b]{0.28\textwidth}
         \centering
         \includegraphics[width=\textwidth]{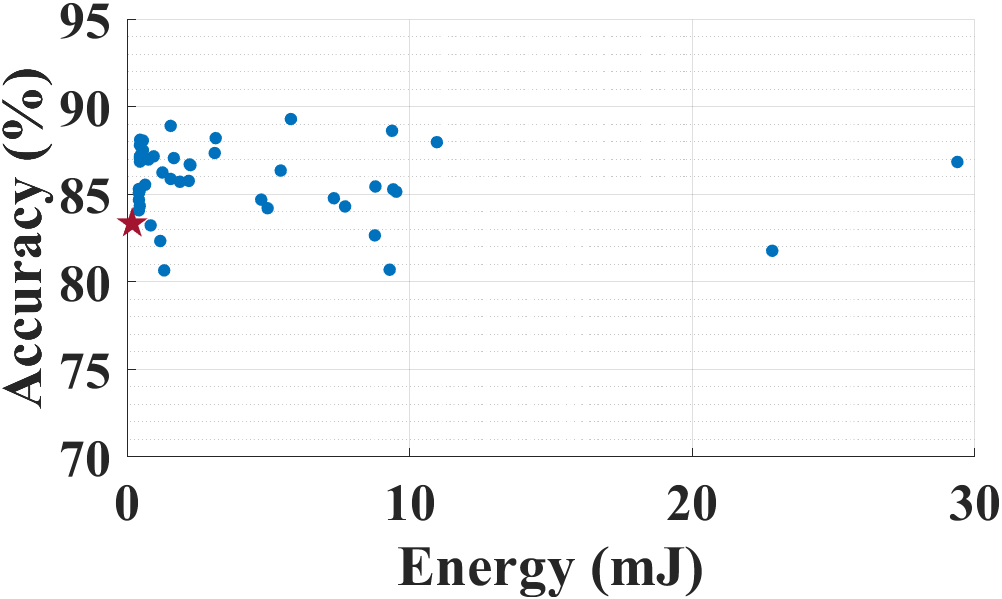}
         \caption{}
         \label{fig:cifar_eng}
     \end{subfigure}
        \caption{Distribution of NAS outputs for CIFAR-10 dataset (The best model is marked by \textcolor{red}{$\star$}) (a) distribution of accuracy when $FF=accuracy$ (b) distribution of accuracy vs latency when $FF=\sfrac{accuracy}{latency}$ (c) distribution of accuracy vs energy when $FF=\sfrac{accuracy}{energy}$.}
        \label{fig:cifar_scat}
\end{figure*}

\begin{table*}[]
\centering
\caption{Best model architecture identified by the proposed NAS method for various datasets and fitness functions. Accuracy, Latency, and Energy metrics are denoted by $Acc$, $Lt$, and $En$, respectively. }
\label{tab:best}
\begin{tabular}{|c|c|cccccc|c|c|c|}
\hline
\multirow{2}{*}{FF} & \multirow{2}{*}{Dataset} & \multicolumn{6}{c|}{Best Model Architecture} & \multirow{2}{*}{Accuracy} & Latency & Energy \\ \cline{3-8}
\parbox[h]{2mm}{\multirow{7}{*}{\rotatebox[origin=c]{90}{$Acc$}}} &  & BT1/K1 & BT2/K2 & BT3/K3 & BT4/K4 & BT5/K5 & BT6/K6 &  & (ms)  & (mJ)  \\ \hline
 & \multicolumn{1}{l|}{} & \multicolumn{1}{l}{} & \multicolumn{1}{l}{} & \multicolumn{1}{l}{} & \multicolumn{1}{l}{} & \multicolumn{1}{l}{} & \multicolumn{1}{l|}{} & \multicolumn{1}{l|}{} & \multicolumn{1}{l|}{} & \multicolumn{1}{l|}{}  \\
\multirow{5}{*}{} & ASL & MVGG/32 & RES/16 & VGG/128 & VGG/128 & RES/128 & RES/64 & 99.98\% & $4.8$ & $1.01$ \\
 & \multicolumn{1}{l|}{} & \multicolumn{1}{l}{} & \multicolumn{1}{l}{} & \multicolumn{1}{l}{} & \multicolumn{1}{l}{} & \multicolumn{1}{l}{} & \multicolumn{1}{l|}{} & \multicolumn{1}{l|}{} & \multicolumn{1}{l|}{} & \multicolumn{1}{l|}{} \\
 & CK+ & RES/16 & VGG/16 & VGG/16 & VGG/64 & MVGG/64 & - & 99.49\% & $4.22$ & $0.273$ \\
 & \multicolumn{1}{l|}{} & \multicolumn{1}{l}{} & \multicolumn{1}{l}{} & \multicolumn{1}{l}{} & \multicolumn{1}{l}{} & \multicolumn{1}{l}{} & \multicolumn{1}{l|}{} & \multicolumn{1}{l|}{} & \multicolumn{1}{l|}{} & \multicolumn{1}{l|}{} \\
 & CIFAR & RES/128 & MVGG/32 & VGG/256 & RES/32 & VGG/128 & RES/256 & 90.24\% & $15$ & $6.7$ \\ \hline
  & \multicolumn{1}{l|}{} & \multicolumn{1}{l}{} & \multicolumn{1}{l}{} & \multicolumn{1}{l}{} & \multicolumn{1}{l}{} & \multicolumn{1}{l}{} & \multicolumn{1}{l|}{} & \multicolumn{1}{l|}{} & \multicolumn{1}{l|}{} & \multicolumn{1}{l|}{}  \\
\multirow{5}{*}{\parbox[h]{2mm}{\multirow{-1.8}{*}{\rotatebox[origin=c]{90}{$Acc/Lt$}}}} & ASL & MVGG/16 & VGG/16 & RES/16 & - & - & - & 99.41\% & $0.78$ & $0.07$ \\
 & \multicolumn{1}{l|}{} & \multicolumn{1}{l}{} & \multicolumn{1}{l}{} & \multicolumn{1}{l}{} & \multicolumn{1}{l}{} & \multicolumn{1}{l}{} & \multicolumn{1}{l|}{} & \multicolumn{1}{l|}{} & \multicolumn{1}{l|}{} & \multicolumn{1}{l|}{} \\
 & CK+ & VGG/16 & VGG/64 & VGG/128 & - & - & - & 97.97\% & $2.1$ & $0.37$ \\
 & \multicolumn{1}{l|}{} & \multicolumn{1}{l}{} & \multicolumn{1}{l}{} & \multicolumn{1}{l}{} & \multicolumn{1}{l}{} & \multicolumn{1}{l}{} & \multicolumn{1}{l|}{} & \multicolumn{1}{l|}{} & \multicolumn{1}{l|}{} & \multicolumn{1}{l|}{}  \\
 & CIFAR & VGG/32 & VGG/32 & RES/32 & - & - & - & 84.07\% & $1.62$ & $0.158$ \\ \hline
  & \multicolumn{1}{l|}{} & \multicolumn{1}{l}{} & \multicolumn{1}{l}{} & \multicolumn{1}{l}{} & \multicolumn{1}{l}{} & \multicolumn{1}{l}{} & \multicolumn{1}{l|}{} & \multicolumn{1}{l|}{} & \multicolumn{1}{l|}{} & \multicolumn{1}{l|}{}  \\
\multirow{5}{*}{\parbox[h]{2mm}{\multirow{-2}{*}{\rotatebox[origin=c]{90}{$Acc/En$}}}} & ASL & VGG/16 & VGG/16 & VGG/64 & - & - & - & 99.27\% & $0.72$ & $0.048$ \\
 & \multicolumn{1}{l|}{} & \multicolumn{1}{l}{} & \multicolumn{1}{l}{} & \multicolumn{1}{l}{} & \multicolumn{1}{l}{} & \multicolumn{1}{l}{} & \multicolumn{1}{l|}{} & \multicolumn{1}{l|}{} & \multicolumn{1}{l|}{} & \multicolumn{1}{l|}{} \\
 & CK+ & VGG/16 & VGG/16 & MVGG/64 & MVGG/16 & - & - & 89.34\% & $2.04$ & $0.15$ \\
 & \multicolumn{1}{l|}{} & \multicolumn{1}{l}{} & \multicolumn{1}{l}{} & \multicolumn{1}{l}{} & \multicolumn{1}{l}{} & \multicolumn{1}{l}{} & \multicolumn{1}{l|}{} & \multicolumn{1}{l|}{} & \multicolumn{1}{l|}{} & \multicolumn{1}{l|}{} \\
 & CIFAR & VGG/32 & VGG/32 & VGG/64 & VGG/128 & VGG/128 & - & 83.33\% & $\num{1.71}$ & $\num{0.183}$ \\ \hline
\end{tabular}
\end{table*}

To validate the effectiveness of our proposed method, we conduct comprehensive experiments utilizing three distinct datasets: (1) the American Sign Language (ASL) Alphabet Dataset \cite{signlangmnistkaggle}, featuring static images for hand gesture classification across 24 letters (excluding motion-dependent J and Z); (2) the Extended Cohn-Kanade (CK+) Dataset \cite{lucey2010extended}, containing 593 video sequences to evaluate facial expression recognition as subjects transition between neutral and seven distinct emotions; and (3) CIFAR-10 \cite{krizhevsky2009learning}, a well-known dataset in the computer vision field for assessing object recognition capabilities with 60,000 color images across 10 classes. To attain the multi-objective optimization, we run our method with three different fitness functions- $accuracy$, $\sfrac{accuracy}{latency}$, and $\sfrac{accuracy}{energy}$, which ensures a holistic view of the method's performance, encompassing both its classification accuracy and operational efficiency across a variety of visual recognition tasks.

Figure \ref{fig:asl_scat}, \ref{fig:ck_scat} and \ref{fig:cifar_scat} show the distribution of NAS outputs for ASL, CK+ and CIFAR-10 datasets, respectively. For the ASL and CK+ datasets, we train 100 distinct models for the optimization of each of the fitness functions and identify the model with the best values of the fitness functions. As the outputs do not have much improvements after around 50 model evaluations, we train 50 models for the CIFAR-10 dataset to save simulation time. Table \ref{tab:best} lists the model descriptions that are identified as the best model in terms of different fitness functions. When the $accuracy$ fitness function is used, the proposed method tends to identify deeper models which leads to higher accuracy due to better generalization in a higher number of layers. The identified models also have a higher number of kernels, which provides a higher number of convolutional filters to allow learning more features. As an instance, for the CIFAR-10 dataset, the best model with the $accuracy$ fitness function picks as many as 6 blocks, which shows $90.24\%$ accuracy, with $15$ ms latency and $6.7$ mJ energy consumption. Two of the blocks have a kernel size of 256, which was set as the highest number of kernels possible for a block in our search space.

With the $\sfrac{accuracy}{latency}$ FF, the proposed method tends to identify shallower models and less number of kernels which leads to less computation cycle and less latency. The energy consumption also reduces at the same time due to less data movement on the IMC chip and the accuracy drops due to fewer features and less generalization of the features. For the CIFAR-10 dataset with $\sfrac{accuracy}{latency}$ fitness function, the identified model uses only three blocks having only 32 kernels each, which shows latency improvement to $1.62$ ms and energy reduction to $0.158$ mJ, but the accuracy drops to $84.07\%$. 

The $\sfrac{accuracy}{energy}$ fitness function tends to identify shallow models and avoids the RES blocks. A possible reason for avoiding the RES blocks might be 
the usage of the global accumulator to implement the skip-connections in the IMC architecture, which leads to more data transfer and higher energy consumption. With $\sfrac{accuracy}{energy}$ fitness function on the CIFAR-10 dataset, a 5-block model is picked where none of the blocks are RES blocks. The model shows as low as $0.183$ mJ energy consumption, with an accuracy drop to $83.33\%$ and a latency of $1.71$ ms. It is noticeable that putting the same weights to accuracy and other hardware metrics leads to significant drops in accuracy in some of the datasets such as CIFAR-10 and CK+ with $\sfrac{accuracy}{energy}$ fitness function. One potential solution is to increase the exponent of the accuracy metric in the fitness function such as $\sfrac{accuracy^n}{energy}$ where $n>1$. We also observe that for none of the datasets and fitness functions, models with 7 or 8 blocks are selected although it was a possibility in the search space. It is possible that those blocks would be used for more challenging datasets with more complex features.

\section{Conclusion}
In this paper, we explored multi-objective neural architecture search for efficient deployment of CNNs in analog in-memory computing (IMC) architectures. We constructed a search space using various building blocks inspired by the convolution layers in VGG and ResNet models. We utilized multi-objective fitness functions combining network accuracy with hardware performance metrics such as latency and energy. Using Bayesian optimization, we explored the search space over three different image classification datasets to identify the best model based on the suggested fitness functions, which revealed some interesting nuances. The exploration with the $accuracy$ fitness function identified deeper models with a higher number of kernels as optimal solutions. When the hardware evaluation metrics are incorporated into the fitness function, such as $\sfrac{accuracy}{latency}$ and $\sfrac{accuracy}{energy}$, shallower models with a lower number of kernels are selected. Moreover, our observations exhibit that the optimization algorithm avoids using RES blocks with skip connection in the network architecture when energy is incorporated into the fitness function. This can be associated with the particular specification of the analog IMC architecture which uses a global accumulator to handle the skip connections.




\section*{Acknowledgment}
This work is supported in part by the National Science Foundation (NSF) under grant number 2409697.



%



\balance
\bibliographystyle{IEEEtran}
\bibliography{refs}

\end{document}